\documentclass[12pt]{article} %
\usepackage[sort]{natbib}
\usepackage{array,epsfig,fancyheadings,rotating}
\usepackage[]{hyperref}  %
\usepackage{sectsty, secdot}
\sectionfont{\fontsize{12}{14pt plus.8pt minus .6pt}\selectfont}
\renewcommand{\theequation}{\thesection\arabic{equation}}
\subsectionfont{\fontsize{12}{14pt plus.8pt minus .6pt}\selectfont}

\usepackage{amsmath}
\usepackage{amssymb}
\usepackage{amsfonts}
\usepackage{multirow}
\usepackage{amsthm}
\usepackage{enumitem}
\usepackage{booktabs}

\theoremstyle{definition}

\usepackage{bm}

\usepackage{hyperref}
\usepackage[capitalise]{cleveref}

\newcommand\logger{{\mathcal L}}
\newcommand\piD{{\mathcal D}}
\newcommand\ts{\textstyle}

\usepackage{shortcuts}

\begin{document}

\renewcommand{\baselinestretch}{2}

\markright{ \hbox{\footnotesize\rm Statistica Sinica
}\hfill\\[-13pt]
\hbox{\footnotesize\rm
}\hfill }

\def\runningtitle{}

\markboth{\hfill{\footnotesize\rm Nathan Kallus} \hfill}
{\hfill {\footnotesize\rm \runningtitle} \hfill}

\renewcommand{\thefootnote}{}
$\ $\par

\fontsize{12}{14pt plus.8pt minus .6pt}\selectfont \vspace{0.8pc}
\centerline{\large\bf Comment: Entropy Learning }
\vspace{2pt} \centerline{\large\bf for Dynamic Treatment Regimes}
\vspace{.25cm} \centerline{Nathan Kallus} \vspace{.25cm} \centerline{\it
Cornell University} \vspace{.55cm} \fontsize{9}{11.5pt plus.8pt minus
.6pt}\selectfont

\def\thefigure{\arabic{figure}}
\def\thetable{\arabic{table}}

\renewcommand{\theequation}{\thesection.\arabic{equation}}

\fontsize{12}{14pt plus.8pt minus .6pt}\selectfont

\setcounter{section}{1} %
\setcounter{equation}{0} %

I would like to congratulate Profs. Binyan Jiang, Rui Song, Jialiang Li, and Donglin Zeng (JSLZ, henceforth) for an exciting development in conducting inferences on optimal dynamic treatment regimes (DTRs) learned via empirical risk minimization using the entropy loss as a surrogate.  JSLZ's ingenuity was to carefully propagate the asymptotic distributions of $M$-estimators through a backward induction using a roll out of estimated individualized treatment regimes (ITRs) learned by weighted entropy loss minimization. This solved an 
open problem on how to conduct rigorous inference on DTRs 
\citep{laber2014dynamic}.

JSLZ's approach leverages
a rejection-and-importance-sampling estimate
of the value of a given decision rule 
based on inverse probability weighting (IPW; see the first unnumbered display equation in JSLZ's Section~2.2)
and its interpretation as a weighted (or cost-sensitive) classification, a celebrated reduction \citep{owl,beygelzimerlangford}. Their use of smooth classification surrogates enables their careful approach to analyzing asymptotic distributions. However, even for evaluation purposes, the IPW estimate is problematic. The estimate is a weighted average of rewards, where, for a horizon of $T$ steps, the weights are the product of $T$ indicators of whether the decision rule's recommendations agree with the observed actions, divided by the product of $T$ propensities for the observed actions. 
With even just two actions per step, the numerator is most often zero. At the same time, the denominator is invariably tiny, and minor differences in probabilities translate into large differences in their inverse products. The result is weights that discard most of the data and are extremely variable on whatever remains. 
This renders the estimator practically useless for any horizon $T$ longer than 2--3 and any reasonably sized sample \citep[see also][]{gottesman2018evaluating}. So, while JSLZ's careful analysis enables us to conduct inferences on DTRs learned by optimizing this estimate (via a surrogate), one might question whether DTRs learned in this way are useful to begin with when $T\geq3$ and $n$ is realistic, given the unreliable evaluation.

\lhead[\footnotesize\thepage\fancyplain{}\leftmark]{}\rhead[]{\fancyplain{}\rightmark\footnotesize\thepage}%

In this comment, I discuss an optimization-based alternative to evaluating ITRs and DTRs, review several connections, and suggest directions forward. 
In \citet{balancepol}, I proposed an approach for evaluating and learning ITRs based on \emph{optimal balance}. Optimal balance -- a technique I have also developed for designing controlled experiments \citep{optapriori}, designing observational studies \citep{gom,matchaistats,deepmatch,komsate}, and estimating marginal structural models \citep{kowmsm} -- directly targets the error objective of interest by optimally choosing weights that minimize it, rather than relying on plug-in-and-pray approaches that fail for practically sized samples, such as IPW. I show how optimal balance extends to DTR evaluation and discuss why it holds promise.

\section*{Balanced Evaluation of ITRs}

JSLZ motivate their approach by first considering ITRs; I will do the same. Indeed, using backward induction, evaluating and learning DTRs reduces to evaluating and learning ITRs. In their Eq.~(2.1), JSLZ recall the central identity of importance sampling, as applied to ITR evaluation, which I repeat here using potential-outcome notation:
\begin{equation}
\label{eq:ipw}\ts
V(\piD\mid X)\equiv{
\Eb{R(a)\int_{a\in\mathcal A}d\piD(a\mid X)\mid X}
}=\Eb{
\frac{\piD(A\mid X)}{\logger(A\mid X)}R
\mid X},
\end{equation}
where
$R(a)$ is the potential reward of action $a$, for any possible action $a\in\mathcal A$ (I make no assumptions on $\mathcal A$; it can be discrete or continuous);
$X\in\mathcal X$ are the prognostic covariates;
$\piD(a\mid X)$ is the probability (usually Dirac) of the decision rule choosing $a$ when seeing $X$;
$A$ and $R$ are the action and reward,  respectively, observed in the data;
$\logger(a\mid X)$ is the probability of $A$, given $X$, in the data;
and we assume ignorable assignment: $R(a)\indep A\mid X\;\forall a\in\mathcal A$.

Given a sample $\{(X_i,A_i,R_i):i\leq n\}$, we can operationalize \cref{eq:ipw} by taking an empirical average of $\frac{\piD(A_i\mid X_i)}{\logger(A_i\mid X_i)}R_i$ (e.g., JSLZ's Eq~(2.3)). However, this can prove problematic in practice, because the density ratio $\frac{\piD(A_i\mid X_i)}{\logger(A_i\mid X_i)}$ can vary wildly, giving some units much higher weight than others and leading to high-variance evaluation. 
Because of this fundamental problem, there have been many variations and iterations of this basic estimator, including weight normalization and clipping \citep{swaminathan2015self}, ``hybrid'' clipping using estimates of $\Eb{R(a)\mid X}$ \citep{tsiatis2007comment,wang2017optimal}, using such estimates as control variates \citep{dudik2011doubly}, optimizing the choice of control variate \citep{cao2009improving,farajtabar2018more}, among others. However, these and other estimators that do not rely completely on extrapolation via outcome modeling need to account for the covariate shift between $\logger$ and $\piD$ and to weight by the density ratio $\frac{\piD(A\mid X)}{\logger(A\mid X)}$, and ultimately suffer from its fundamental instability. 
This is particularly problematic when $\piD(A\mid X)$ is Dirac, as is usually the case since optimal policies are deterministic, because it means that any data point that disagrees with $\mathcal D$'s recommendation is discarded, even if informative.
Smoothing $\piD(A\mid X)$ amounts to shrinking the estimate, by linearity.
(When $A$ is continuous, this means \emph{all} data points are discarded; smoothing, as in \citealp{continuoustreat}, becomes a necessity.)

I briefly explain my optimal balancing proposal for ITR evaluation from \citet{balancepol}.
Given \emph{any} outcome-weighted estimator, $\hat V=\frac1n\sum_{i\leq n}W_iR_i$, with $W=W(X_{1:n},A_{1:n})$, its conditional mean squared error, given the data upon which the weights depend, decomposes to:
$$\ts
\Eb{\prns{\hat V-\frac1n\sum_{i\leq n}V(\piD\mid X_i)}^2\mid X_{1:n},A_{1:n}}=B^2(\mu;W)+\frac1{n^2}\sum_{i\leq n}W_i\sigma_i^2,
$$
where $\sigma_i^2=\op{Var}\prns{R_i\mid X_i,A_i}$, $\mu(x,a)=\Efb{R_i\mid X_i=x,A_i=a}$, and
$$\ts
B(f;W)=\frac1n\sum_{i\leq n}\int_{a\in\mathcal A}f(X_i,a)d(W_i\delta(a-A_i)-\piD(a\mid X_i)),
$$
which, for every $W$, is a linear operator on the space of functions $[\mathcal A\times\mathcal X\to\Rl]$. (A similar result holds if we augment the weighted estimator with an estimate $\hat\mu$, as in AIPW.) Because $\mu$ (or the difference $\mu-\hat\mu$) is unknown, this suggests seeking weights $W$ that make $B(f;W)$ small for many functions $f\in\mathcal F$. Under appropriate conditions, 
$$\ts\sup_{f\in\mathcal F}B(f;W)=\sup_{\|f\|\leq 1}B(f;W)=\|B(\;\cdot\;;W)\|_*,
$$
where $\|\cdot\|$ is the gauge of $\mathcal F$ and $\|\cdot\|_*$ its dual. 
Thus, we seek weights $W$ that make the norm of the operator $B(\;\cdot\;;W)$ small, subject to some 2-norm regularization in order to control the variance.
Because setting $W_i=\frac{\piD(A_i\mid X_i)}{\logger(A_i\mid X_i)}$ makes $B(f;W)$ a sum of independent mean-zero terms, a straightforward empirical process argument \citep[see, e.g.,][]{pollard} shows that, under appropriate conditions on $\mathcal F$, these weights also make $\|B(\;\cdot\;;W)\|_*\to0$. However, in practice, these plug-in weights still have all the problems of extreme values and being mostly zeros. Instead, my proposal for optimally balanced evaluation of ITRs is to choose weights that directly optimize the error objective of interest:
\begin{equation}\label{eq:baloptcrosssec}\ts
W^*\in\argmin_{W\geq0\;:\;\frac1n\sum_{i\leq n}W_i=1}\ \|B(\;\cdot\;;W)\|_*^2+\frac{\lambda}{n^2}\|W\|_2^2,
\end{equation}
which is a linearly constrained convex optimization problem.

\newcommand{\directM}{-0.114}
\newcommand{\directS}{0.308}
\newcommand{\directE}{0.328}
\newcommand{\IPWregtruAM}{-0.005}
\newcommand{\IPWregtruAS}{2.209}
\newcommand{\IPWregtruAE}{2.209}
\newcommand{\IPWregestAM}{-0.491}
\newcommand{\IPWregestAS}{0.310}
\newcommand{\IPWregestAE}{0.581}
\newcommand{\IPWregtruBM}{-0.402}
\newcommand{\IPWregtruBS}{0.329}
\newcommand{\IPWregtruBE}{0.520}
\newcommand{\IPWregestBM}{-0.514}
\newcommand{\IPWregestBS}{0.242}
\newcommand{\IPWregestBE}{0.568}
\newcommand{\IPWnormtruAM}{-0.181}
\newcommand{\IPWnormtruAS}{0.487}
\newcommand{\IPWnormtruAE}{0.519}
\newcommand{\IPWnormestAM}{-0.250}
\newcommand{\IPWnormestAS}{0.415}
\newcommand{\IPWnormestAE}{0.485}
\newcommand{\IPWnormtruBM}{-0.211}
\newcommand{\IPWnormtruBS}{0.405}
\newcommand{\IPWnormtruBE}{0.457}
\newcommand{\IPWnormestBM}{-0.251}
\newcommand{\IPWnormestBS}{0.390}
\newcommand{\IPWnormestBE}{0.463}
\newcommand{\DRregtruAM}{0.435}
\newcommand{\DRregtruAS}{4.174}
\newcommand{\DRregtruAE}{4.196}
\newcommand{\DRregestAM}{0.259}
\newcommand{\DRregestAS}{0.451}
\newcommand{\DRregestAE}{0.520}
\newcommand{\DRregtruBM}{0.267}
\newcommand{\DRregtruBS}{0.432}
\newcommand{\DRregtruBE}{0.508}
\newcommand{\DRregestBM}{0.230}
\newcommand{\DRregestBS}{0.361}
\newcommand{\DRregestBE}{0.428}
\newcommand{\DRnormtruAM}{0.408}
\newcommand{\DRnormtruAS}{0.634}
\newcommand{\DRnormtruAE}{0.754}
\newcommand{\DRnormestAM}{0.471}
\newcommand{\DRnormestAS}{0.550}
\newcommand{\DRnormestAE}{0.724}
\newcommand{\DRnormtruBM}{0.428}
\newcommand{\DRnormtruBS}{0.544}
\newcommand{\DRnormtruBE}{0.692}
\newcommand{\DRnormestBM}{0.467}
\newcommand{\DRnormestBS}{0.511}
\newcommand{\DRnormestBE}{0.692}
\newcommand{\optIPWAM}{0.227}
\newcommand{\optIPWAS}{0.163}
\newcommand{\optIPWAE}{0.280}
\newcommand{\optDRAM}{-0.006}
\newcommand{\optDRAS}{0.251}
\newcommand{\optDRAE}{0.251}
\newcommand{\IPWsupportM}{13.6}
\newcommand{\IPWsupportS}{2.9}
\newcommand{\optsupportM}{90.7}
\newcommand{\optsupportS}{3.2}

\begin{table}[t!]\scriptsize%
\def\baselinestretch{1}\selectfont%
\caption{ITR evaluation performance in \citet[Example~1]{balancepol}}\label{tableex1}%
\centering%
\begin{tabular}{lp{0pt}crrp{0pt}crrp{0pt}r}\toprule
\multicolumn{1}{c}{\multirow{2}{*}{{\centering Weights}}}
&& \multicolumn{3}{c}{Outcome Weighting} && \multicolumn{3}{c}{Augmented OW (DR)}
&& \multicolumn{1}{c}{\multirow{2}{*}{
$\fmagd W_0$%
}}\\[-.3em]\cmidrule{3-5}\cmidrule{7-9}
&& \multicolumn{1}{c}{RMSE}& \multicolumn{1}{c}{Bias} & \multicolumn{1}{c}{SD} 
&& \multicolumn{1}{c}{RMSE}& \multicolumn{1}{c}{Bias} & \multicolumn{1}{c}{SD}  
\\[-.2em]\midrule
IPW%
&& $\IPWregtruAE$ & $\IPWregtruAM$ & $\IPWregtruAS$
&& $\DRregtruAE$  & $\DRregtruAM$  & $\DRregtruAS$
&& $\IPWsupportM\pm\IPWsupportS$
\\
NIPW%
&& $\IPWnormtruAE$ & $\IPWnormtruAM$ & $\IPWnormtruAS$
&& $\DRnormtruAE$  & $\DRnormtruAM$  & $\DRnormtruAS$
&& $\IPWsupportM\pm\IPWsupportS$
\\[-.2em]
\cmidrule{1-11}
Balanced
&& $\mathbf{\optIPWAE}$ & $\optIPWAM$ & $\optIPWAS$
&& $\mathbf{\optDRAE}$  & $\optDRAM$  & $\optDRAS$
&& $\optsupportM\pm\optsupportS$
\\\bottomrule
\end{tabular}
\end{table}

To illustrate how this works, I include an excerpt from \citet{balancepol} in \cref{tableex1}, where I apply this to an example with $\abs{\mathcal A}=5$, $n=100$, and low overlap between $\logger$ and $\piD$. For simplicity, I let $\mathcal F$ be the unit ball of the RKHS with kernel $\mathcal K((x,a),(x',a'))=\delta(a-a')e^{-\|x-x'\|_2^2}$ and $\lambda=1$. I include augmented (DR) estimators, using $\hat\mu$ fitted by XGBoost, as well as normalized (H\'ajek) IPW. IPW discards about 86\% of the data; the balanced approach only 9\%, and correspondingly performs much better.

\section*{Balanced Evaluation of DTRs}

When considering sequential decisions, the fragility of IPW only becomes worse: the weights become even sparser and more extreme, because they are now the ratio of the product of $T$ indicators and the product of $T$ probabilities. Fortunately, the approach to balanced evaluation extends to the case of DTRs, which holds promise for salvaging DTR value estimators that rely on density ratio weighting in any way.

In the sequential setting, 
we are interested in evaluating the DTR value:
\begin{align*}
V(\piD_{1:T})&\equiv\ts\sum_{t\leq T}
\braces{V_t(\piD_{1:t})\equiv\ts\E{
\int_{ a_{1:t}\in {\mathcal A}_{1:t}}
{R_t( a_{1:t})
d\piD_{1:t}(a_{1:t}\mid  X_{1:t}( a_{1:t-1}), a_{1:t-1})
}
}}
,
\end{align*}
where 
$\piD_{1:t}(a_{1:t}\mid  X_{1:t}( a_{1:t-1}), a_{1:t-1})=
\prod_{s\leq t}\piD_s(a_s\mid  X_{1:s}( a_{1:s-1}), a_{1:s-1})$
and, for each $t$ and sequence of actions $a_{1:t}\in{\mathcal A_{1:t}}=\mathcal A_1\times\cdots\times\mathcal A_t$, we now have potential outcomes for both the reward at time $t$ and the time-dependent covariates at time $t+1$. 
Our data consist of observations of trajectories $ X_{1:T}, A_{1:T}, R_{1:T}$, assuming
sequentially ignorable assignment:
$$
{R_{t:T}( a_{1:T}),X_{t+1:T}( a_{1:T-1})}
\indep
A_t( a_{1:t-1})
\mid
 X_{1:t}( a_{1:t-1}),
 A_{1:t-1}( a_{1:t-2}).
$$
As in the case of ITRs, consider estimating $V_t(\piD_{1:t})$ by a weighted average of outcomes. To streamline the already cumbersome notation, I discuss this in terms of population averages. Thus, I consider the weighted average of observables $\hat V_t=\Efb{W_{1:t}R_t}$, for some weights $W_{1:t}=\prod_{s\leq t}W_s$ where $W_s=W_s(X_{1:s}, A_{1:s})$. Then, iteratively applying sequential ignorability yields a decomposition similar to the ITR case:
\begin{align}
&\label{eq:dtrdecom}\ts\hat V_t-V_t(\piD_{1:t})=
\sum_{s\leq t}B_{s}(\mu_{t,s};W_s),\\\notag
&\ts B_{s}(f;W_s)\equiv
\E{
\int_{a_s\in\mathcal A_s}
f\prns{ X_{1:s}, A_{1:s-1},a_s}
d\prns{W_s\delta\prns{a_s-A_s}-
\piD_s(a_s\mid X_{1:s}, A_{1:s-1})
}
},
\\\notag
&\ts \mu_{t,s}( x_{1:s}, a_{1:s})\equiv
W_{1:s-1}(x_{1:s-1},a_{1:s-1})
\Eb{
R^{\piD}_{t,s}(a_{1:s})
\mid
X_{1:s}= x_{1:s},
 A_{1:s-1}= a_{1:s-1}
}
,\\\notag
&\ts R^{\piD}_{t,s}( a_{1:s})\equiv
\int_{a_{s+1:t}\in\mathcal A_{s+1:t}}
R_t( a_{1:t})
d\piD_{s+1:t}(a_{s+1:t}\mid X_{1:t}(a_{1:t-1}), a_{1:t-1}).
\end{align}
This looks rather complicated, but has a simple message: the error is a sum over $s=1,\dots,t$ of a particular moment mismatch ($B_s$) in variables $X_{1:s},A_{1:s}$ between the weighted data distribution and the distribution induced by deviating and following $\piD_s$ at step $s$. Therefore, to obtain a good estimate, we require weights that make this mismatch small for many functions $f:{\mathcal X}_{1:s}\times{\mathcal A}_{1:s}\to\Rl$. As before, setting $W_s=\frac{\piD_s(A_s\mid  X_{1:s}, A_{1:s-1})}{\logger_s(A_s\mid  X_{1:s}, A_{1:s-1})}$ achieves this at the population level or for very large samples, but can fail horribly in realistically sized samples. (JSLZ actually use weights $\prod_{s=1}^T\frac{\piD_s(A_s\mid  X_{1:s}, A_{1:s-1})}{\logger_s(A_s\mid  X_{1:s}, A_{1:s-1})}$ on $\sum_{t\leq T}R_t$, which is also unbiased, but even more unstable; when estimating the average reward at time $t$, multiplying by density ratios for times after $t$ is superfluous and just increases the variance.) However, given any sample and some function class $\mathcal F_s$, we can seek weights that minimize the (empirical) worst-case mismatches $\|B_{s}(\;\cdot\;;W_s)\|_{s*}$, subject to some 2-norm regularization to control the variance. Doing so amounts to nothing more than solving \cref{eq:baloptcrosssec}, for each of $t=1,\dots,T$, to obtain $W_t$, each time considering $X_{1:t},A_{1:t-1}$ as the ``prognostic covariates'' being balanced and $a_t$ as the ``action.'' 
(We could have also placed the $W_{1:s-1}$ term in $B_s$, rather than in $\mu_{t,s}$, which would have amounted to a simple reweighting of the moment conditions being balanced; however, I focus on the simplest reduction to repeatedly solving problems of the form of \cref{eq:baloptcrosssec}. We can also apply \cref{eq:dtrdecom} to the residuals and use an augmented DR-style estimator.)

\begin{table}[t!]\scriptsize%
\def\baselinestretch{1}\selectfont%
\caption{DTR evaluation performance}\label{tableex2}%
\centering%
\begin{tabular}{lp{0pt}rrrp{0pt}rrrp{0pt}rrr}\toprule
\multicolumn{1}{c}{\multirow{2}{*}{{\centering Weights}}}
&&
\multicolumn{3}{c}{$T=3$} && 
\multicolumn{3}{c}{$T=5$} && 
\multicolumn{3}{c}{$T=7$}
\\[-.3em]\cmidrule{3-5}\cmidrule{7-9}\cmidrule{11-13}
&& \multicolumn{1}{c}{RMSE}& \multicolumn{1}{c}{Bias} & \multicolumn{1}{c}{SD} 
&& \multicolumn{1}{c}{RMSE}& \multicolumn{1}{c}{Bias} & \multicolumn{1}{c}{SD} 
&& \multicolumn{1}{c}{RMSE}& \multicolumn{1}{c}{Bias} & \multicolumn{1}{c}{SD}  
\\[-.2em]\midrule
IPW$_T$&&$5e2$&$0.96$&$5e2$&&$4e4$&$-42.94$&$4e4$&&$2e2$&$28.61$&$2e2$\\
IPW&&$2e2$&$0.41$&$2e2$&&$1e4$&$-11.52$&$1e4$&&$1e4$&$-2.08$&$1e4$\\
NIPW$_T$&&$11.82$&$8.39$&$8.32$&&$38.07$&$38.01$&$2.03$&&$63.10$&$63.09$&$0.64$\\
NIPW&&$6.90$&$4.64$&$5.10$&&$26.94$&$26.27$&$5.96$&&$51.57$&$51.22$&$5.98$
\\[-.2em]
\cmidrule{1-13}
Bal. $\mathcal K_G$&&$\mathbf{6.28}$&$-0.57$&$6.26$&&$\mathbf{11.73}$&$9.69$&$6.61$&&$\mathbf{18.65}$&$17.44$&$6.61$\\
Bal. $\mathcal K_M$&&$6.87$&$-0.26$&$6.87$&&$12.71$&$10.06$&$7.78$&&$19.43$&$17.80$&$7.78$\\
\bottomrule
\end{tabular}
\end{table}
\section*{A DTR Evaluation Example}

To demonstrate how this works, I include a simple example.
Let $T$ vary and, for $t\leq T$, let
$\mathcal A_t=\{-1,+1\}$, 
$\mathcal X_t=\R 2$, 
$R_t(a_{1:t})=5a_t+X_{t,1}(a_{t-1})+\epsilon_t$, 
$\epsilon_t\sim\mathcal N(0,1)$,
$X_{1,j}\sim\mathcal N(0,1)$,
$X_{t+1,j}(a_{1:t})=a_t+X_{t,j}(a_{t-1})+\xi_{t,j}$, 
$\xi_{t,j}\sim\mathcal N(0,1)$,
$\logger(+1\mid x_{1:t},a_{1:t-1})=\op{expit}(2(X_{t,1}+X_{t,2})A_{t-1})$, and
$\piD(+1\mid x_{1:t},a_{1:t-1})=\indic{(X_{t,1}+X_{t,2})A_{t-1}<0}$.
I consider 2,000 replications of $n=800$ for each $T\in\{3,5,7\}$.
To apply balanced evaluation, I let $\mathcal F_t$ be the unit ball of the RKHS 
with kernel 
$\mathcal K((x_{1:t},a_{1:t}),(x'_{1:t},a'_{1:t}))=\delta(a_{t-1:t}-a'_{t-1:t})\mathcal K_x(x_{t},x'_{t})$, where $\mathcal K_x$ is either the Gaussian ($\mathcal K_G$) or Mat\'ern ($\mathcal K_M$, $\nu=5/2$) kernel.
I compare this with IPW and normalized IPW. I also include the variation in JSLZ 
in which we multiply $\sum_{t\leq T}R_t$ by density ratios up to $T$, referred to as IPW$_T$.

The results appear in \cref{tableex2}.
The large variance of IPW renders it unusable even with a reasonably sized data set. The variance is so large that it throws off the bias estimated by 2,000 replications (zero in theory). NIPW mitigates this variance, but is actually equal to the uniform weights 37\%, 99\%, or 100\% of the time, for $T=3,5,7$, respectively, and has correspondingly large bias. Balancing has both low bias (indistinguishable from that estimated for IPW) and low variance (comparable to NIPW).

Estimating DTR value when horizons are long is a fundamentally difficult task. Whereas IPW discards most of the data, estimating reward and transition models requires strong modeling assumptions and precarious extrapolations. Balancing could provide a fruitful middle ground: rather than throwing away imperfectly matching trajectories, we imbue the problem with some structure to allow these to be used, while ensuring that our weights achieve the same consistency guarantees afforded by IPW asymptotically (see, e.g., \citealp{balancepol,gom}).

\section*{Beyond Evaluation: Learning and Inference}

I have argued the merits of using optimal balance to evaluate DTRs. An immediate question is how to use this to learn DTRs. As before, we can optimize the value estimate.
Although computationally challenging, this is the approach I took in \citet{balancepol} for ITRs. To apply this to DTRs requires just an application of backward induction with roll out.

With regard to inference (JSLZ's primary concern),
this remains open for the balanced approach, but there may be promising directions.
Asymptotically, under appropriate conditions on $\mathcal F$ and the class of rules being considered, optimal sample weights will uniformly concentrate, so we may consider the distribution when we use the optimal population weights. However, it remains unclear how the estimated rules are distributed (even ITRs). A possible hybrid approach is to use JSLZ's Eq.~(2.8), but to replace $\prod_{s\geq{t+1}}\frac{\piD_s(A_s\mid  X_{1:s}, A_{1:s-1})}{\logger_s(A_s\mid  X_{1:s}, A_{1:s-1})}$ with the optimal balancing weights $W^*_{t+1:T}$, while keeping $\frac{\piD_t(A_t\mid  X_{1:t}, A_{1:t-1})}{\logger_t(A_t\mid  X_{1:t}, A_{1:t-1})}$ and replacing its numerator with a smooth surrogate. This will at least alleviate issues with longer horizons by limiting IPW to one step, while still being an $M$-estimator.

While JSLZ's advance is a breakthrough, further advances are necessary.
Currently, using IPW and its derivatives to evaluate and learn DTRs when $T$ is moderate and $n$ is realistic is woefully impractical.

\markboth{\hfill{\footnotesize\rm Nathan Kallus} \hfill}
{\hfill {\footnotesize\rm \runningtitle} \hfill}

\bibhang=1.7pc
\bibsep=2pt
\fontsize{9}{14pt plus.8pt minus .6pt}\selectfont
\renewcommand\bibname{\large \bf References}
\expandafter\ifx\csname
natexlab\endcsname\relax\def\natexlab#1{#1}\fi
\expandafter\ifx\csname url\endcsname\relax
  \def\url#1{\texttt{#1}}\fi
\expandafter\ifx\csname urlprefix\endcsname\relax\def\urlprefix{URL}\fi

\bibliographystyle{chicago}      %
\bibliography{comment}   %

\vskip .65cm
\noindent
School of Operations Research and Information Engineering and Cornell Tech, Cornell University, New York, NY 10044, USA.
\vskip 2pt
\noindent
kallus@cornell.edu

\end{document}